\documentclass[lettersize,journal]{IEEEtran}
\usepackage{amsmath,amsfonts}
\usepackage{array}
\usepackage{textcomp}
\usepackage{stfloats}
\usepackage{url}
\usepackage{verbatim}
\usepackage{graphicx}
\usepackage{cite}
\usepackage{pifont}
\usepackage{epsfig,rotating,setspace,latexsym,amsmath,epsf,amssymb,amsfonts,bm,theorem,cite,algorithm,graphicx,epsf,authblk,epstopdf,color,algpseudocode,bbm}
\newcounter{procedure}
\newcounter{algorithm saved}
\usepackage{algorithm} 
\usepackage{algpseudocode} 
\algnewcommand\algorithmicforeach{\textbf{for each:}}
\algnewcommand\ForEach{\item[ \algorithmicforeach]}
\usepackage{caption}
\usepackage{subfigure}
\usepackage{subcaption}
\usepackage{caption}
\usepackage{multirow} 

\makeatletter
\newenvironment{procedure}[1][htb]{%
    \renewcommand{\ALG@name}{Procedure}
    \setcounter{algorithm saved}{\value{algorithm}} 
    \setcounter{algorithm}{\value{procedure}}
    \begin{algorithm}[#1]%
    }{\end{algorithm}
    \setcounter{procedure}{\value{algorithm}}
    \setcounter{algorithm}{\value{algorithm saved}}
}
\usepackage{amsmath}

\usepackage{tikz}
\usetikzlibrary{arrows.meta,positioning,fit,calc,shapes.misc}
\begin{document}

\title{\texttt{FB-NLL}: A Feature-Based Approach to Tackle Noisy Labels in Personalized Federated Learning}

\author{Abdulmoneam~Ali,~\IEEEmembership{Graduate Student Member,~IEEE,}
        and~Ahmed~Arafa,~\IEEEmembership{Member,~IEEE,}
\thanks{The authors are with the Department
of Electrical and Computer Engineering, University of North Carolina at Charlotte,
NC 28223, USA. e-mails: aali28@charlotte.edu, aarafa@charlotte.edu.}
\thanks{This work was supported by the U.S. National Science Foundation under Grants CNS 21-14537 and ECCS 21-46099, and has been presented in part at the Asilomar Conference on Signals, Systems, and Computers, Pacific Grove, CA, USA, October 2024 \cite{ds_asilomar24}, and at the IEEE International Conference on Communications (ICC), Montreal, Canada, June 2025 \cite{RCC_icc25}.}}

\maketitle

\begin{abstract}
Personalized Federated Learning (PFL) aims to learn multiple task-specific models rather than a single global model across heterogeneous data distributions. Existing PFL approaches typically rely on iterative optimization—such as model update trajectories—to cluster users that need to accomplish the same tasks together. However, these learning-dynamics-based methods are inherently vulnerable to low-quality data and noisy labels, as corrupted updates distort clustering decisions and degrade personalization performance. To tackle this, we propose \texttt{FB-NLL}, a feature-centric framework that decouples user clustering from iterative training dynamics. By exploiting the intrinsic heterogeneity of local feature spaces, \texttt{FB-NLL} characterizes each user through the spectral structure of the covariances of their feature representations and leverages subspace similarity to identify task-consistent user groupings. This geometry-aware clustering is label-agnostic and is performed in a one-shot manner prior to training, significantly reducing communication overhead and computational costs compared to iterative baselines.

Complementing this, we introduce a feature-consistency-based detection and correction strategy to address noisy labels within clusters. By leveraging directional alignment in the learned feature space and assigning labels based on class-specific feature subspaces, our method mitigates corrupted supervision without requiring estimation of stochastic noise transition matrices. 
In addition, \texttt{FB-NLL} is model-independent and integrates seamlessly with existing noise-robust training techniques. Extensive experiments across diverse datasets and noise regimes demonstrate that our framework consistently outperforms state-of-the-art baselines in terms of average accuracy and performance stability.

\end{abstract}


\section{Introduction}

Federated learning (FL) is a privacy-preserving machine learning paradigm that enables users to collaboratively train a model without sharing their local datasets \cite{pmlr-v54-mcmahan17a}. A key advantage of FL is that learning occurs closer to where the data is stored. However, while keeping data on user devices strengthens privacy guarantees, it also introduces two fundamental challenges: ensuring reliability and quality of local data, and addressing heterogeneity in data distributions among users. 

Ensuring data quality is particularly challenging in FL since it varies from user to user depending on how they collect their data, and there is no central node that has access to all collected training data that can monitor its quality. This makes the FL setting more vulnerable to noisy data, and has motivated a substantial body of research aimed at developing methods to ensure data quality in a distributed manner while minimizing the communication and computational overhead and preserving users' privacy. 

Another major challenge is data heterogeneity, meaning that users' data follow different distributions. This implies the potential need for multiple global models, each capable of fitting different users' datasets. Consequently, this has motivated the change in the standard FL setting from learning a universal model that fits all user data points towards personalized models tailored to each user\cite{towards_pfl}. Such customized models have been shown to accelerate the convergence of learning and have given rise to the personalized federated learning (PFL) setting. 

In PFL, users with similar objectives collaborate to learn a common model. However, how to effectively cluster users with \textit{similar} objectives, while maintaining their privacy, so that they can learn jointly remains challenging in PFL--a problem commonly referred to in the literature as the \textit{cluster-identity estimation} problem. The most widely adopted approach to address this challenge is based on the behavior of each user's training samples with respect to each personalized model. Specifically, users are clustered according to the model that minimizes their loss function \cite{clustered_FL, over_air_cfl}. The central idea is that similar model behavior on a pair of users' training data indicates that they may be targeting the same underlying learning objective. We review related work on the clustering problem in PFL in Section~\ref{subsec:cluster_related}.

This model-based clustering approach, however, overlooks the earlier challenge of reliability and quality of the local datasets at each user. In particular, it implicitly assumes that the loss function is evaluated on {\it clean} training samples. In practice, users may have {\it noisy} labeled data samples obtained, e.g., from low quality crowd-sourcing platforms, since high quality datasets are often difficult to acquire \cite{noisy_label_sources}. The noisy label problem is more pronounced in FL because user-side data may be collected from diverse sources, each with its own stochastic noise. Thus, assuming high-quality labels for training data is overly optimistic, and relying on the loss function to solve the cluster-identity problem needs to be reconsidered. This gives rise to our first question that we are interested to address: 
\begin{center}
\textit{Can we solve the cluster-identity estimation problem for PFL without relying on the loss function or estimating the noise model for each user?}
\end{center} 
We answer the above question in the affirmative. In our proposed clustering solution, we focus on studying the behavior of clustering users to learn multiple models in settings with significant label noise. We show that, even without explicit data cleaning or noise-robust algorithms, a {\it one-shot data similarity-based clustering} method can learn from data corrupted by arbitrary level of label noise. The key idea is that the proposed data-similarity algorithm is label-agnostic, relying on features rather than labels; consequently, the cluster-identity estimation problem can be effectively resolved. 

Although our clustering-based algorithm is robust to label noise, the issue of noisy labels cannot be overlooked due to their detrimental effect on the learning process. Training in the presence of noisy labels often causes neural networks to memorize the noise, which in turn degrades generalization performance \cite{liu2020early}. A rich body of literature proposes a variety of approaches to address the problem of training with noisy labels in FL, which we summarize in Section~\ref{subsec:FNLL}. 

Most existing solutions are iterative, depend on the learning dynamics of individual samples, and remain tightly coupled to the training process. This motivates our second research question:
\begin{center}
\textit{After clustering users, can we reduce each user's noise level before training begins?}
\end{center} 
We also answer this question in the affirmative. By leveraging structure in the feature space, we can infer class-level relationships and identify mislabeled samples prior to training, thereby reducing the effective noise ratio and improving the downstream learning performance.

Motivated by these insights, we propose a new PFL framework that jointly addresses cluster-identity estimation and noisy-label detection and correction, while requiring minimal communication and no training iterations.
\begin{figure*}[t]
    \centering
    \includegraphics[width=1\linewidth]{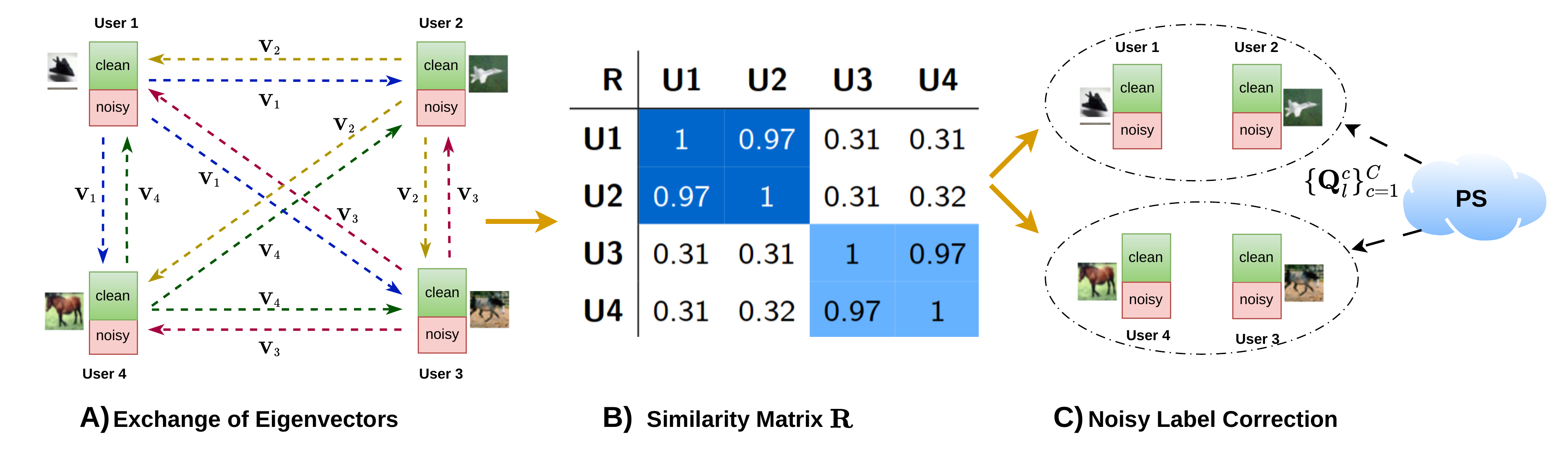}
    \caption{Proposed feature-centric framework. In (\textbf{A}), users exchange covariance matrix eigenvectors, $\{\mathbf{V}_i\}_{i=1}^{4}$, to quantify cross-user distribution alignment. (\textbf{B}) The Parameter Server (PS) constructs a similarity matrix $\mathbf{R}$ to identify distinct user clusters. (\textbf{C}) The PS provides clean spectral references, $\{\mathbf{Q}_l^c\}_{c=1}^{C}$, to facilitate relabeling the users' noisy local data.} 
    \label{fig:overview}
\end{figure*}

\subsection{Contributions}

We investigate whether cluster-identity estimation and noisy-label correction in PFL can be addressed directly through users’ data distributions in the feature space, without relying on gradients, weight updates, or training dynamics. Our framework leverages the \textit{spectral} structure of covariance matrices to construct compact representations of data distributions, enabling user clustering in a fully feature-centric manner. Similarly, to identify and correct noisy labels, we exploit the geometric properties of the feature space: candidate noisy samples are projected onto class-specific subspaces, which allows us to detect and reassign mislabeled points based on their alignment with these subspaces after projection.
 
To ensure practical applicability in federated environments, we design a communication-efficient algorithm in which users exchange only a small number of eigenvectors of their feature covariance matrices, which is significantly lighter than exchanging model weights or gradients. The overall framework leads to two orthogonal algorithmic components---one for clustering and one for noisy-label correction---that can operate independently or be seamlessly combined. An overview of our proposed approach is depicted in Fig.~\ref{fig:overview}.

Our main contributions are summarized as follows:

\begin{itemize}
\item \textbf{Cluster-identity estimation}: We address the problem of clustering users in PFL, where users aim to learn different tasks. In particular, we propose a method that measures the alignment between users' data distributions in a privacy-preserving manner. This one-shot data similarity method relies only on spectral information extracted from the features of each user's local data.

\item \textbf{Noisy-label detection}: We extend data similarity to the class level within each local dataset. For each local class, the global parameter server and the users exchange low-dimensional spectral signatures and compute a symmetric similarity score between the local class and each global clean class; classes with high mutual similarity are then automatically relabeled at the class level, enabling efficient detection of fully mislabeled classes.

\item \textbf{Noisy-label correction}: For classes that cannot be relabeled as a whole due to spectral misalignment with all global clean classes, we perform fine-grained, per-sample correction. Each sample is projected onto the principal subspaces of all classes and reassigned to the class whose subspace yields the largest projection, thereby correcting partially noisy labels at the sample level.

\item \textbf{Algorithmic properties}: Our framework is one-shot, training-free, communication-efficient, model-agnostic, and modular, making it suitable for a wide range of PFL settings.

\item \textbf{Empirical validation}: We demonstrate that our approach outperforms state-of-the-art FL baselines in both accuracy and variance reduction. For a fair comparison, we evaluate the clustering component separately from the noisy-label correction component. In addition, we extensively test the framework under diverse noise regimes, including task-level structured noise (class-dependent and class-independent) and user-level noise (uniform-noise). These scenarios span both homogeneous setting, where all users share the same noise level, and heterogeneous settings characterized by varying fractions of noisy users and user-specific noise levels.

\end{itemize}

\subsection{Related work}

\subsubsection{Clustering in PFL}\label{subsec:cluster_related} Most approaches for addressing cluster-identity estimation in PFL are iterative and learning-based.
In these methods, user associations are inferred from information extracted during the training trajectory, such as loss values, model weights, or gradients. For example, \cite{clf} uses cosine similarity between users' weight updates as a proxy for similarity between their underlying data distributions. Similarly, \cite{Many-taskFL} computes cosine similarity between layer-wise network weights for pairs of users and uses these scores for clustering. Alternatively, \cite{clustered_FL} proposes an alternating minimization framework that iteratively estimates cluster memberships and updates local models. Specifically, each user selects the model that minimizes its loss, adopts the corresponding cluster identity, and updates its local model accordingly, after which a new personalized model is computed by aggregating the models of users within the same cluster.
In \cite{NIPS2017}, the authors assume the existence of a matrix capturing relationships between tasks. This matrix, either known beforehand or estimated during the training, is incorporated into the objective function together with a regularization term that depends on the task-relation matrix. The training process then proceeds with the proposed augmented objective function.  %
Overall, these model-based clustering strategies are iterative and dynamic; cluster memberships may change from one communication round to another as the model parameters evolve.

\subsubsection{Federated Noisy Label Detection and Correction} \label{subsec:FNLL}

Existing approaches in this area of machine learning can be divided into three main categories: detecting samples with noisy labels; refining the loss function to improve noise robustness; or designing specialized training strategies. 

Methods for noisy-label detection often focus on estimating the stochastic transition matrix that maps clean labels to their noisy counterparts~\cite{trans_estimation,li2022estimating}. To avoid the difficulty of estimating the transition matrix, the authors of \cite{yue2024ctrl} identify noisy samples by analyzing the training-loss trajectory of each data point. Based on this trajectory, each sample is then classified as having either a clean or noisy label. Approaches based on loss-function refinement aim to mitigate overfitting to noisy labels by incorporating regularization terms or by designing noise-tolerant loss functions~\cite{patrini2017making, wei2023mitigating}. The third line of work involves training an auxiliary network for sample weighting or supervision-level filtering \cite{jiang2018mentornet}, \cite{han2018coteaching}. 

While centralized noisy-label learning techniques can, in principle, be adapted to the federated setting, their effectiveness is fundamentally limited by inherent challenges of FL such as heterogeneous, user-dependent noise patterns and strict privacy constraints. Consequently, new methods specifically tailored to FL have been developed. The authors of \cite{fang2022robust, RHFL_tmc} introduce a reweighting strategy that reduces the influence of noisy users by downweighting their contributions during global aggregation.

An alternative strategy is to partition users into clean and noisy groups, and allow noisy users to utilize the updated model trained on clean users to relabel their corrupted data.
This approach is employed in \cite{FedCorr} and \cite{wu2023fednoro}. The key difference between these two frameworks lies in how clean and noisy users are identified: \cite{FedCorr} uses the local intrinsic dimension as the distinguishing criterion, whereas \cite{wu2023fednoro} relies on their training losses.
Alternatively, \cite{ClientPruning} proposes a user-pruning mechanism that excludes potentially noisy users entirely, where the clean/noisy decision is made by evaluating each user’s loss on a clean public dataset at the parameter server. Rather than discarding or downweighting noisy users, \cite{FedNed} leverages them through negative distillation, compelling the global model to diverge from the incorrect predictions produced by these users. 

In scenarios where all users possess noisy labels, \cite{RoFL} proposes a mechanism that exchanges class-wise feature centroids between users and the server to align class representations and enforce consistent decision boundaries. In \cite{FedLSR}, the authors develop an implicit regularization strategy that mitigates overfitting to noisy labels by mixing the predictions of original and augmented samples and then sharpening the mixed predictions to increase the model confidence.



We summarize prior work addressing the noisy-label challenge in FL in Table \ref{tab:FNLL_survey}. We categorize the evaluation criteria as follows: C1 indicates algorithms that perform noise detection; C2 corresponds to methods that adopt noise correction; C3 denotes approaches that are model-independent; and C4 highlights methods that are communication-efficient.
\begin{table}[h]
\captionsetup{labelfont=normalfont,textfont=normalfont}
\centering
\begin{tabular}{|c|c|c|c|c|c|} 
\hline
Algorithm & C1 & C2 & C3 & C4 \\
\hline
FedCorr~\cite{FedCorr} & \ding{51} & \ding{51}  & \ding{55} & \ding{55}  \\
ClipFL~\cite{ClientPruning} & \ding{51} & \ding{55} & \ding{55}  & \ding{55}    \\
FedNoRo~\cite{wu2023fednoro} & \ding{51} & \ding{51} & \ding{55}& \ding{51} \\

RoFL~\cite{RoFL} & \ding{51} & \ding{51} & \ding{55} & \ding{55}   \\
FedNed~\cite{FedNed} &  \ding{51}  & \ding{51} & \ding{55}  & \ding{55}   \\
RHFL~\cite{RHFL_tmc} & \ding{51}    & \ding{51} & \ding{55}  & \ding{55}    \\
FedLSR~\cite{FedLSR}  & \ding{51}    & \ding{51} & \ding{55}  & \ding{51}   \\
 FB-NLL (ours) & \ding{51}   & \ding{51}   & \ding{51}  & \ding{51}  \\
\hline
\end{tabular}
\caption{Comparison of representative federated noisy-label learning algorithms across four criteria: noisy-label detection (C1), noisy-label correction (C2), model-agnostic design (C3), and communication efficiency (C4).}
\label{tab:FNLL_survey}
\end{table}

\textbf{Notation and Organization.} We use bold uppercase letters to denote matrices, bold lowercase letters to denote vectors, and non-bold symbols to denote scalars. For a positive integer $C$, we write $[C] \triangleq \{1,2,\dots,C\}$. For a vector $\mathbf{x}$, $\|\mathbf{x}\|_2$ denotes its $\ell_2$ norm. The remainder of the paper is organized as follows. Section~\ref{sec:sys_model} presents the system model and the label noise models. Section~\ref{sec:data-sim_clustering} introduces the proposed data-similarity clustering algorithm, while Section~\ref{sec:noise_correction} presents the proposed noise-correction algorithm. Section~\ref{sec:experiments} presents the experimental setup and results. Section~\ref{sec:conclusion} concludes the paper.

\section{System Model and Label Noise}\label{sec:sys_model}

We consider a FL system consisting of a parameter server (PS) and a set of users $\mathcal{K}=\{1,2,\dots,K\}$. Each user aims to learn a specific task drawn from a total of $M$ different tasks. All tasks are defined over the same underlying dataset. Specifically, given a dataset with class labels $\{1,\dots, C\}$, each task corresponds to a subset of these labels, and the tasks are mutually disjoint. That is, the $m$th task $T_m \subseteq \{1,\dots, C\}$, $\forall m \in [M]$, with $T_m \cap T_{\bar{m}}= \emptyset$, $\forall m \neq \bar{m}$. We denote the set of all tasks as $\mathcal{T}\triangleq\{T_1,\dots,T_M\}$. Since each user is interested in learning a specific task $\tau_k \in \mathcal{T}$, the goal of the system is not to learn a single global model, but rather to learn $M$ personalized models---one per task---to meet the users' requirements.

Each user $k \in \mathcal{K}$ has a local dataset $D_{k}=\{(x_{(k,i)},y_{(k,i)})\}_{i=1}^{n_k}$, where $x_{(k,i)}\in\mathbb{R}^p$ is the $i$th $p$-dimensional feature vector, $y_{(k,i)} \in \{1,\dots, C\}$ is the corresponding label, and $n_k$ is the number of training samples at user $k$. We assume a label-skewed non-IID distribution, where the majority of a user’s samples belong to its intended task $\tau_k$, while a minority originate from unintended tasks in $\mathcal{T} \setminus \tau_k$. 
In addition, we assume that all users are honest and trustworthy. Given that multiple users may share an interest in the same task, it is reasonable to cluster them together.

The main goal is to learn $M$ models by clustering users, without violating their privacy, such that each cluster contains only users aiming to learn the same task. 

We assume that users' samples are not entirely clean and that some labels are noisy. We consider two main approaches to model the unknown stochasticity of these noisy samples, which we illustrate next.

\subsection{Task-Dependent Noise Model}

In this noise model, we assume that each user $k$ has a fraction $\alpha n_{k}$ of samples with corrupted labels, for some $\alpha\in(0,1)$. We design this noise model to approximate a worst-case corruption pattern that significantly challenges the clustering procedure. 
Specifically, the label-flipping mechanism is {\it asymmetric} across the label space: all randomly selected true labels associated with a given task are mapped to a single noisy label belonging to a different (unintended) task.
This concentrated cross-task corruption increases ambiguity during clustering, which motivates the term \textit{task-dependent noise}. We note that this setting can be viewed as a variation of the structure-biased noise described in \cite{rolnick1705deep}, where clean labels are mapped to multiple noisy labels with different probabilities, whereas in our case the corruption is concentrated toward a single unintended label.

Building on this setting, we further consider two sub-models of task-dependent noise: {\it class-independent} and {\it class-dependent} noise. Both models share the property that the labels selected for corruption are mapped to a single noisy label chosen {\it uniformly} at random from classes belonging to unintended tasks. However, they differ in how the samples subject to corruption are selected, as detailed below.

In the class-independent noise model, the samples selected for corruption are chosen {\it independently} of their true labels. Specifically, for user $k$, $\alpha n_k$ samples are drawn uniformly at random from its entire local dataset and then reassigned to a label selected uniformly at random from an unintended task, i.e., from $\mathcal{T}\setminus \tau_k$. In contrast, under the class-dependent noise model, the corrupted samples originate from a {\it specific} class within the intended task. Concretely, for user $k$, one label is first selected uniformly at random from $\tau_k$, and then $\alpha n_k$ samples belonging to that label are all flipped to a single label chosen from $\mathcal{T}\setminus \tau_k$.

\begin{figure*}[t] 
    \centering
    \subfigure[Data with clean labels]{%
        \includegraphics[width=0.244\linewidth]{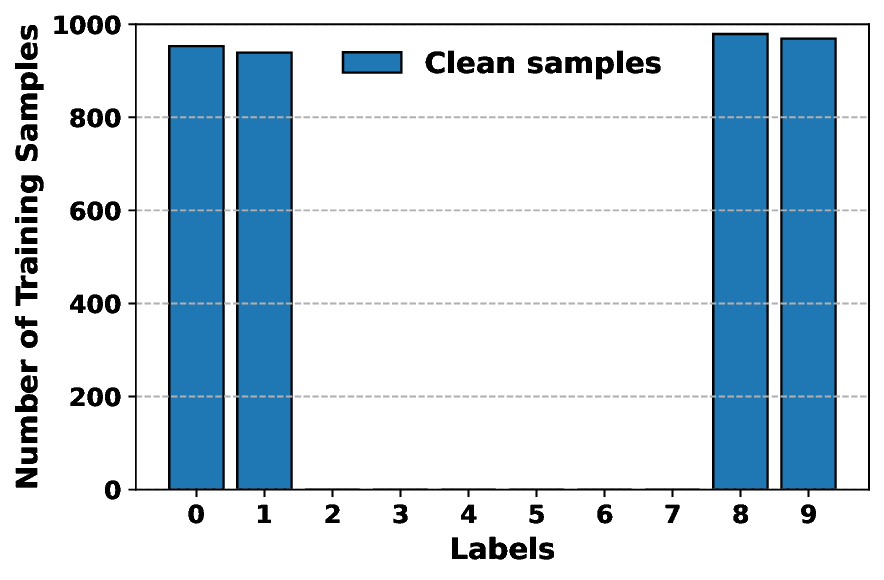}%
        \label{fig:a}%
    }%
    \hfill%
    \subfigure[Data under class-independent noisy labels]{%
        \includegraphics[width=0.244\linewidth]{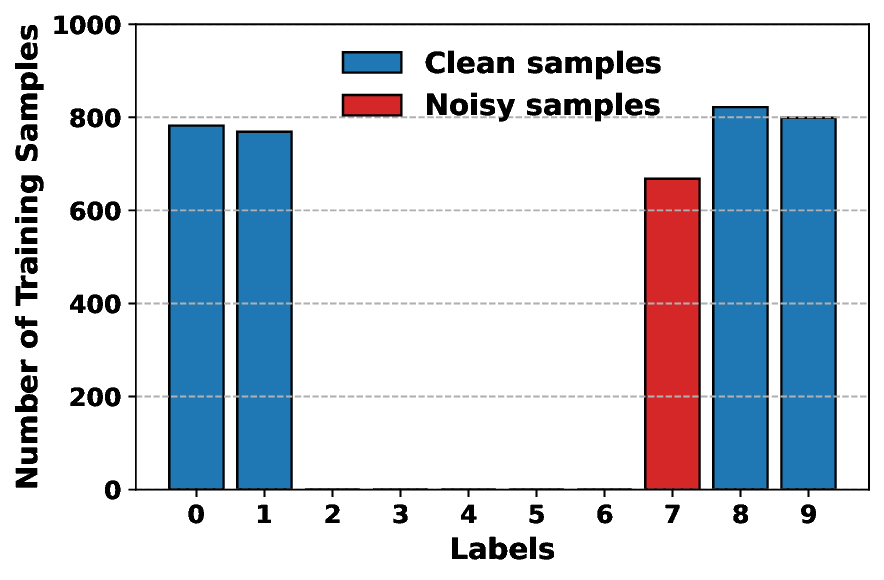}
        \label{fig:b}%
    }%
    \hfill%
    \subfigure[Data under class-dependent noisy labels]{%
        \includegraphics[width=0.244\linewidth]{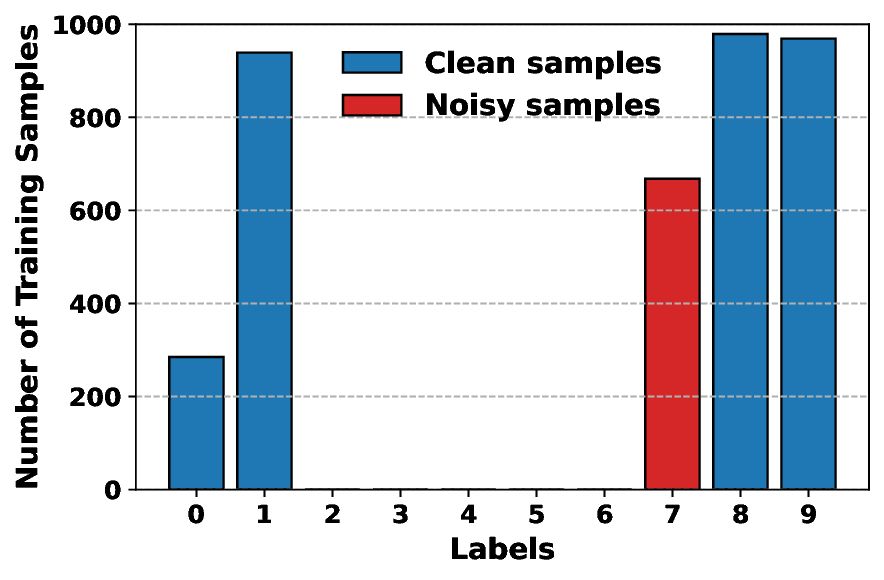}%
        \label{fig:c}%
    }%
    \hfill%
    \subfigure[Data under uniform noisy labels]{%
        \includegraphics[width=0.244\linewidth]{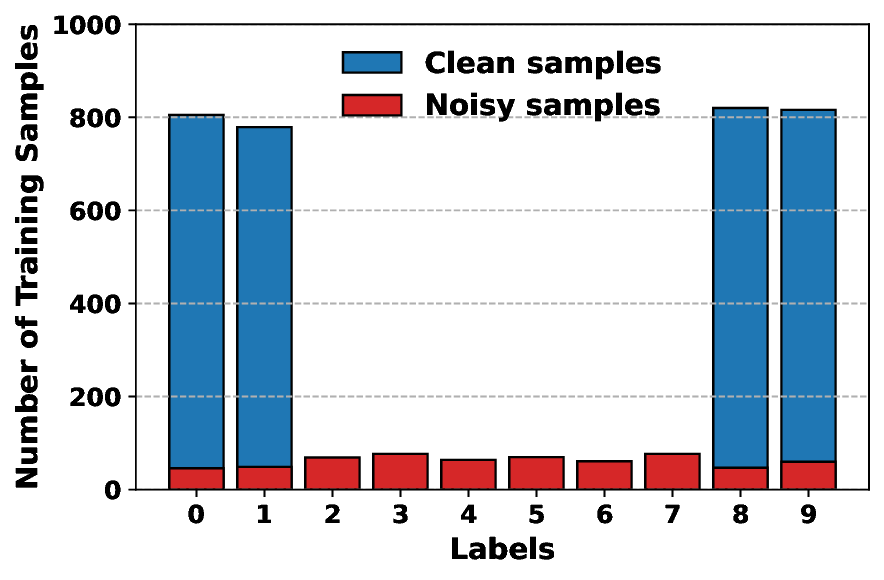}%
        \label{fig:d}%
    }%
    \caption{
    Visualization of the three noise models applied to a CIFAR-10 user whose intended task is to classify non-living object classes $\{0,1,8,9\}$: (a) Clean labels, (b) Class-independent noise, (c) Class-dependent noise, and (d) Uniform noise. The noise rate is $\alpha = 0.17$, with zero labels from unintended tasks.}
    \label{fig:different_noisy_models}
\end{figure*}

Hence, the main distinction between the two models lies in the source of the corrupted samples. In the class-independent model, the selected samples may originate from multiple class labels, whereas in the class-dependent model all corrupted samples are drawn from a specific class in $\tau_k$. In the class-dependent setting, if the number of samples in the selected source class is fewer than $\alpha n_k$, the remaining samples required to reach $\alpha n_k$ are drawn from another class in $\tau_k$. Consequently, a user may end up with one fully corrupted class and one or more partially corrupted classes. It is important to note that sampling is performed without replacement. Therefore, unlike the structure-biased noise model in \cite{rolnick1705deep}, no clean duplicates of corrupted samples remain in the dataset. 

The rationale behind these two sub-models of task-dependent noise is grounded in practical data-collection scenarios, where a user's dataset may originate from multiple sources, each exhibiting distinct corruption patterns. In the class-independent model, each data source contributes proportionally to the corrupted samples, resulting in noise that is uniformly distributed across the user’s local dataset. While under the class-dependent model, corruption is concentrated within a specific source (i.e., a particular class), and the remaining sources provide clean data. 
\subsection{Task-Independent (Uniform) Noise}

Under this noise model, the fraction of corrupted samples at user $k$ is denoted by $\alpha_k$. In contrast to the previously described task-dependent models, the noise levels $\{\alpha_k\}$ may vary across users and may be zero for some $k$, implying that not all users necessarily have noisy data and that certain users may possess entirely clean local datasets. 

For noisy users, samples are corrupted in a task-agnostic (uniform) manner, in contrast to the structured noise models described earlier, and hence the name \textit{task-independent noise}. Specifically, noisy samples are selected uniformly at random from the user's local dataset, and their labels are replaced with labels drawn uniformly from the $C$ classes. This model is widely adopted in the noisy-label learning literature; see, e.g., \cite{FedCorr, TrustBCFL, li2024feddiv}. Formally, each user $k \in [K]$ is independently designated as noisy with probability $\rho \in [0,1]$. Conditional on user $k$ being noisy, its local noise level $\alpha_k$ is drawn independently from the uniform distribution over $(\beta, 1)$, where $\beta>0$ denotes the minimum noise severity. Clean users have zero noise. Hence, the noise level at user $k$ is defined as 
\begin{align}
    \alpha_k =
    \begin{cases}
        u, \quad u \sim \mathcal{U}(\beta, 1), & \text{with probability } \rho, \\
        0, & \text{with probability } 1 - \rho,
    \end{cases}
\end{align}
where $\mathcal{U}(\beta,1)$ denotes the uniform distribution over $(\beta,1)$. This model captures heterogeneous noise severity across users without imposing any task-dependent structure on the corruption process. 

In our setting, we apply a single noise model across all users. That is, we do not consider the hybrid settings in which some users are affected by task-independent noise while others are subject to task-dependent noise.\footnote{However, our proposed framework is amenable to such scenarios, as both the clustering and relabeling procedures are agnostic to specific noise patterns.}

To further elucidate the behavior of the three noise models, Fig.~\ref{fig:different_noisy_models} presents a representative realization for a user whose local training samples are drawn from the CIFAR-10 dataset. The intended task for this user is the classification of non-living objects, corresponding to labels $\{0, 1, 8, 9\}$. Fig.~\ref{fig:a} shows the clean local raining data with ground-truth labels. In this realization, all samples belong to the intended task. Fig.~\ref{fig:b} illustrates the class-independent noise model with noise rate $\alpha = 0.17$. Under this model, an equal proportion of samples from each class is corrupted (with high probability). Accordingly, approximately $\frac{1}{4} \times 17\%$ of the samples from each of the user’s classes are reassigned to label $7$ (chosen at random from $\{2,3,4,5,6,7\}$). Fig.~\ref{fig:c} depicts the class-dependent noise model. In this realization, class~$0$ is randomly selected as the corrupted class, and its corrupted samples are reassigned to label $7$. Finally, Fig.~\ref{fig:d} presents the uniform noise model. With parameter $\beta=0.1$, the user-specific noise level satisfies $\alpha \in (\beta,1)$ (here, $\alpha=0.17$). Each sample is independently corrupted with probability $\alpha$; conditional on corruption, its label is mapped to a label drawn uniformly at random from the $C$ classes.

Next, we discuss our proposed clustering algorithm.

\section{Data Similarity PFL Clustering} \label{sec:data-sim_clustering}

To enable collaborative learning among users targeting the same personalized model, they should be assigned to the same cluster. Specifically, users seeking to learn the $m$th task should be grouped together to learn the corresponding parameter $w_{m}$.
Toward that end, each user $k$ in the $m$th cluster runs stochastic gradient descent (SGD) and shares its updated weight $w_{(k,m)}^{(t)}$ with the PS during global communication round $t$. The PS then uses FedAvg \cite{pmlr-v54-mcmahan17a} to send $w_m^{(t+1)}$ back to cluster $m$. The main challenge in applying the FedAvg algorithm among users with the same personalized model is that their cluster identities are unknown. To efficiently estimate users' cluster identities, we adopt a data valuation-based technique originally proposed in \cite{data_valuation}, and later extended in our preliminary works to the multi-task federated learning setting  \cite{ds_asilomar24} and the noisy label setting \cite{RCC_icc25}. The key idea is to estimate dataset similarity by comparing how different datasets vary along principal components learned from other datasets. This similarity-based approach underlies our answer to the first question about estimating cluster identities in a training-free manner. In this work, we further develop this idea and show that the resulting  data similarity algorithm is robust in realistic and challenging scenarios, and outperforms several baselines.

The data similarity algorithm consists of three main steps. In the first step, each user $k$ applies a spectral decomposition of its local feature covariance matrix:
\begin{align}\label{eq:eigens}
\frac{1}{n_k}\Phi(\mathbf{X}_k)^{\top}\Phi(\mathbf{X}_k) = \mathbf{V}_k \operatorname{diag}(\bm{\lambda}_k) \mathbf{V}_k^\top,
 \end{align}
 where $\mathbf{X}_k \in \mathbb{R}^{n_k\times p}$ denotes the data matrix of user $k$, and $\Phi(\cdot) \in \mathbb{R}^{n_k\times d}$ represents the corresponding feature representations with $d < p$. The vector $\bm{\lambda}_k \in \mathbb{R}^d$ contains the eigenvalues, and $\mathbf{V}_k \in \mathbb{R}^{d \times d}$ contains the associated eigenvectors.

In the second step, users exchange their top-$q$ eigenvectors to enable cross-user projection and similarity estimation. Notably, this exchange preserves data privacy since neither raw data nor exact covariance statistics are shared.\footnote{The true eigenvalues are not transmitted; only eigenvector information is exchanged.} 
The cross-user spectral response is obtained by projecting the eigenvectors of user $j$ onto user $k$'s feature covariance and evaluating the Euclidean norm to get the following ``energy'' term: 
\begin{align}\label{eq:eigen_proj}
    e_{i}^{(k,\, j)}=\left\|\frac{1}{n_k}\Phi(\mathbf{X}_k)^{\top}\Phi(\mathbf{X}_k) \, \mathbf{v}_{i}^{(j)}\right\|_2, \quad \forall i \in [q], 
\end{align}
where $\mathbf{v}_{i}^{(j)}$ is the $i$-th eigenvector of user $j$. The quantity $e_{i}^{(k,j)}$ represents the spectral response of user $k$'s covariance along the $i$-th principal direction of user $j$. It is worth noting that if $\mathbf{v}_i^{(j)}$ coincides with an eigenvector $\mathbf{v}_i^{(k)}$ of user $k$ associated with eigenvalue $\lambda_i^{(k)}$, then we would have
\[
 e_{i}^{(k,\, j)} = \lambda_i^{(k)},
\]
since $\frac{1}{n_k}\Phi(\mathbf{X}_k)^{\top}\Phi(\mathbf{X}_k)$ is positive semidefinite and $\|\mathbf{v}_i^{(k)}\|=1$. Hence, the proposed quantity recovers the true eigenvalue when the eigen-directions of the two users are aligned. Deviations from this value reflect misalignment between their principal subspaces.

Based on the local eigenvalues and the cross-user spectral responses, each user computes a {\it relevance/similarity} metric defined as:
\begin{align}
   s_{i}^{(k,j)}=& \frac{\min\{\lambda_i^{(k)},e_i^{(k,\,j)}\}} {\max\{\lambda_i^{(k)},e_i^{(k,\,j)}\}}, \quad \forall i \in [q], \label{eq:lambd(i,j)}\\
    r(k,\,j)=&\left(\prod_{i=1}^{q}  s_{i}^{(k,\,j)}\right)^{\frac{1}{q}}.\label{eq:r(i,j)}
\end{align}
The normalization via the $\min/\max$ ratio ensures that each term lies in $[0,1]$ and equals one if and only if the projected energy matches the local eigenvalue along that direction. We assume that the eigenvalues of each user are ordered in descending magnitude, and the similarity metric is computed using the top-$q$ principal components, ensuring consistent spectral importance across users.

The intuition behind \eqref{eq:lambd(i,j)} and \eqref{eq:r(i,j)} is that eigenvalues characterize how the variance (or information content) is distributed across principal directions in $\mathbb{R}^d$. If user $j$'s principal directions align well with those of user $k$, then the projected energy $e_i^{(k,j)}$ will closely match $\lambda_i^{(k)}$, resulting in $s_i^{(k,j)} \approx 1$ for all $i$ and hence $r(k,j) \approx 1$. Conversely, significant subspace misalignment reduces the projected energy and decreases the overall similarity score. Moreover, the geometric mean aggregation in \eqref{eq:r(i,j)} ensures that the similarity metric is sensitive to misalignments across the entire subspace. Unlike the arithmetic mean, which may be dominated by a single well-aligned principal component, the geometric mean effectively penalizes the overall score if any principal direction exhibits poor alignment, thereby providing a more robust measure of true distributional overlap.


Finally, in the third step, users transmit their relevance scores, defined in \eqref{eq:r(i,j)}, to the PS. Since the relevance metric is generally asymmetric, i.e., we may have $r(k,j) \neq r(j,k)$, the PS computes a symmetric average similarity as follows:
\begin{align}\label{eq:avg_r}
    [\mathbf{R}]_{k,j}=\frac{r(k,j)+r(j,k)}{2}, \quad \forall k, j \in \mathcal{K}.
\end{align}
By construction, $\mathbf{R}\in\mathbb{R}^{K\times K}$ is  symmetric and serves as the pairwise similarity matrix. The PS then applies Hierarchical Agglomerative Clustering (HAC) \cite{pml1Book} on $\mathbf{R}$ to determine each user's cluster assignment. The resulting cluster identities are encoded in a matrix $\bm{CI} \in \{0,1\}^{K\times M}$, where $\bm{CI}[k,m]=1$ indicates that user $k$ belongs to cluster $m$. Each user is assigned to exactly one cluster, satisfying 
\[
\sum_{m=1}^{M} \bm{CI}[k,m] = 1, \quad \forall k \in \mathcal{K}.
\]

The detailed similarity-based clustering procedure is summarized in Steps 2--14 of Algorithm~\ref{alg:main_algo}.

To better illustrate the proposed similarity measure, we present an example of the matrix $\mathbf{R}$ in Table~\ref{tab:rel_matrix}. In this example, the CIFAR-10 dataset is distributed among five users such that users $1$ and $2$ have samples from the classes $\{$plane, car, ship, truck$\}$, while users $3, 4,$ and $5$ have samples from the classes $\{$bird, cat, deer, dog, frog, horse$\}$. As observed from Table~\ref{tab:rel_matrix}, users $1$ and $2$ exhibit high mutual similarity, while users $3, 4,$ and $5$ form another highly similar group. In contrast, the similarities across the two groups are significantly smaller, clearly revealing two underlying task clusters.

\begin{table}[t]
\captionsetup{labelfont=normalfont,textfont=normalfont}
\centering
\begin{tabular}{c|ccccc}
\hline
${\mathbf{R}}$ & User $1$ & User $2$ & User $3$ & User $4$ & User $5$ \\
\hline
User $1$ & $1$ & $0.97$ & $0.31$ & $0.31$ & $0.32$ \\
User $2$ & $0.97$ & $1$ & $0.31$ & $0.32$ & $0.32$ \\
User $3$ & $0.31$ & $0.31$ & $1$ & $0.97$ & $0.98$ \\
User $4$ & $0.31$ & $0.32$ & $0.97$ & $1$ & $0.98$ \\
User $5$ & $0.32$ & $0.32$ & $0.98$ & $0.98$ & $1$\\
\hline
\end{tabular}
\caption{Example of the data similarity matrix $\mathbf{R}$ on the CIFAR-10 dataset with two underlying tasks.}
\label{tab:rel_matrix}
\end{table}

The HAC algorithm initially assigns each user to an individual cluster and iteratively merges the most similar clusters according to the similarity matrix $\mathbf{R}$. In this example, users $1$ and $2$ merge first due to their high similarity, as do users $4$ and $5$. Subsequently, user $3$ joins the cluster formed by users $4$ and $5$. The clustering process can be visualized via a dendrogram, where different cutting heights correspond to different numbers of clusters. 
Since this example involves two underlying tasks, the dendrogram is cut to produce $M = 2$ clusters. In general, the HAC algorithm is configured to produce $M$ clusters corresponding to the number of personalized models to be learned.

After determining the cluster identities, the PS initiates the training phase by broadcasting the initial model weights $\{w_{m}^{(0)}\}_{m=1}^M$, one for each cluster. Each user $k$ assigned to cluster $m$ performs local SGD for $E$ epochs and sends its updated weight $w_{(k,m)}^{(t)}$ to the PS at the $t$-th communication round. Subsequently, the PS aggregates the model parameters within each cluster independently, as detailed in Steps 16--25 of Algorithm~\ref{alg:main_algo}. 

We refer to the clustering-only variant of the proposed method in this work as {\it Feature-Based Noisy Label Learning} (denoted \texttt{FB-NLL(-)}), which corresponds to Algorithm~\ref{alg:main_algo} without the noise detection and correction step.\footnote{Algorithm~\ref{alg:main_algo} includes an additional noise detection and correction step (line 15), which is described in the next section. The clustering and training components, however, can be executed independently without this step.}

\begin{procedure}[t]
\caption{Similarity Criterion}\label{proc:data_relevance_general}
\begin{algorithmic}[1]

\State \textbf{Input}: data matrix $\mathbf{X}$, received eigenvectors $\mathbf{W}=[\mathbf{w}_1,\dots,\mathbf{w}_q]$, truncation rank $q$

\Statex \underline{\textbf{(Local spectral signature)}}
\If{$(\bm{\lambda},\mathbf{V},\bm{\Sigma})$ not cached for $\mathbf{X}$}
    \State $n \gets$ number of samples in $\mathbf{X}$
    \State $\bm{\Sigma} \gets \frac{1}{n}\Phi(\mathbf{X})^\top \Phi(\mathbf{X})$
    \State $(\bm{\lambda},\mathbf{V}) \gets \text{TopEig}(\bm{\Sigma}, q)$
    \Comment{$\bm{\lambda}\in\mathbb{R}^q,\ \mathbf{V}\in\mathbb{R}^{d\times q}$}
\EndIf

\Statex \underline{\textbf{(Cross-user spectral response)}}
\For{$i \in [q]$}
    \State $e_i \gets \left\| \bm{\Sigma}\mathbf{w}_i \right\|_2$
\EndFor

\Statex \underline{\textbf{(Directional relevance and aggregation)}}
\For{$i \in [q]$}
    \State $s_i \gets \dfrac{\min\{\lambda_i, e_i\}}{\max\{\lambda_i, e_i\}}$ \Comment{$\lambda_i$ denotes the $i$th entry of $\bm{\lambda}$}
\EndFor

\State $r \gets \left(\prod_{i=1}^{q} s_i\right)^{1/q}$

\State \textbf{Output}: $r,\ (\bm{\lambda},\mathbf{V})$

\end{algorithmic}
\end{procedure}


The proposed clustering algorithm relies solely on feature representations, as indicated in \eqref{eq:eigens}. Since clustering decisions are derived from the spectral structure of the feature covariance matrices, they are independent of label information and therefore inherently robust to noisy labels. In contrast, many existing approaches perform clustering based on training dynamics, such as model updates or gradient similarity, which can be significantly distorted in the presence of label noise.

From a communication perspective, the proposed method requires exchanging only the top-$q$ eigenvector matrix $\mathbf{V} \in \mathbb{R}^{d \times q}$ once during the clustering phase. Weight-based clustering schemes, on the other hand, typically rely on iterative model exchanges and repeated optimization steps to jointly determine cluster identities and update model parameters. The feature-based clustering step, however, is performed in a {\it one-shot manner prior to training}, substantially reducing communication overhead during cluster identification. 
Given these properties, our proposed \texttt{FB-NLL} framework is both communication-efficient and robust to noisy supervision, making it well suited for heterogeneous and challenging federated settings. 

Next, we build upon the clustering framework to detect and correct noisy labels within each identified cluster (also prior to training).

\section{Noisy Label Detection and Correction}\label{sec:noise_correction}

Motivated by the effectiveness of the feature-based clustering approach in identifying users with similar task distributions, we extend the same spectral similarity principle to the class level to detect and correct noisy labels within each identified cluster. This class-wise extension yields a fully feature-centric PFL framework that not only clusters users accurately but also mitigates label noise prior to training, thereby enhancing robustness and improving downstream model performance.

To enable users within each cluster to identify and correct potentially noisy labels in their local data, we assume that the PS has access to a relatively small, clean labeled validation dataset. This assumption is well-established in recent robust FL and PFL \cite{ClientPruning, FedNed,RHFL_tmc, martin_jaggi_kd, li2023communicationefficient}. Notably, this dataset is typically minimal in size and disjoint from the users’ local data, ensuring it serves only as a geometric reference. 

For each class $c \in [C]$, the PS maintains $n_s^c$ clean feature vectors stacked into a matrix $\mathbf{X}_s^c \in \mathbb{R}^{n_s^c \times d}$. The PS computes the class-specific covariance matrix 
\begin{align*}
    \bm{\Sigma}_s^c = \frac{1}{n_s^c} \Phi(\mathbf{X}_s^c)^\top \Phi(\mathbf{X}_s^c),
\end{align*}
and performs an eigen-decomposition 
\begin{align*}
    \bm{\Sigma}_s^c = \mathbf{Q}^c \operatorname{diag}(\bm{\lambda}_s^c) (\mathbf{Q}^c)^\top.
\end{align*}
The PS retains the top-$l$ eigenvectors $\mathbf{Q}_l^c \in \mathbb{R}^{d \times l}$ corresponding to the largest eigenvalues. These class-specific principal subspaces are used for noisy label detection and correction over two main phases as detailed below.

\textbf{Phase 1: Class-Wise Noise Detection and Correction.} In this phase, the PS shares the class-specific principal subspaces $\mathbf{Q}_l^c$ with the users. Each user performs a {\it class-wise} eigen-decomposition (as in \eqref{eq:eigens}) to obtain the eigenvectors $\mathbf{V}^{\tilde{c}}$ corresponding to its (potentially noisy) class $\tilde{c}$ data, and uploads them to the PS.\footnote{That is, the user constructs its class $\tilde{c}$ feature covariance matrix $\frac{1}{n^{\tilde{c}}} \Phi(\mathbf{X}^{\tilde{c}})^\top \Phi(\mathbf{X}^{\tilde{c}})$, applies \eqref{eq:eigens} to get the eigenvectors matrix $\mathbf{V}^{\tilde{c}}$. We drop the user index here for ease of presentation.} The PS then measures the alignment between each clean class $c$ and the received eigenvectors using the data similarity framework introduced in Section~\ref{sec:data-sim_clustering}. Specifically, it applies equations \eqref{eq:eigen_proj}, \eqref{eq:lambd(i,j)} and \eqref{eq:r(i,j)} using the clean feature matrix $\mathbf{X}_s^c$ and the eigenvectors in $\mathbf{V}^{\tilde{c}}$. 
The PS then returns the set of similarity scores $\{r(c,\tilde{c})\}_{c=1}^C$ to the user. 

Simultaneously, the user computes its own similarity score using the PS's clean eigenvectors $\mathbf{Q}_l^c$ following the same similarity procedure, and averages the two directional similarity scores for each class. 

In case the averaged similarity score between the user’s class $\tilde{c}$ and a {\it unique} clean class $c$ exceeds a predefined threshold $\tau_{\mathrm{sim}}$ (i.e., only one clean class satisfies the threshold), the entire class $\tilde{c}$ is relabeled as $c$. Otherwise, if no clean class satisfies the threshold or multiple clean classes do, a deeper investigation is conducted in Phase~2.

\textbf{Phase 2: Sample-Wise Noise Detection and Correction.} This phase is only employed in case Phase 1 is indefinite. Using the class-specific principal subspaces $\mathbf{Q}_l^c$ obtained in Phase~1, we proceed with sample-wise relabeling.

Let $\mathbf{z}$ denote the feature representation of a sample whose label is potentially noisy (and whose class did not satisfy the conditions in Phase~1). Inspired by \cite{kim2021fine}, we solve the problem of estimating the true label of $\mathbf{z}$ by identifying the class subspace that best represents it. Specifically, the feature vector $\mathbf{z}$ is projected onto the principal subspace of each class $c \in [C]$, and its label is updated according to
\begin{align}
    \tilde{c} \leftarrow \arg\max_{c \in [C] } \left\|\mathbf{Q}_l^c  (\mathbf{Q}_l^c)^{\top}\, \mathbf{z}\right\|_2. 
\end{align}
That is, the sample is assigned to the class whose principal subspace yields the largest projection magnitude. 

The two phases of identifying and relabeling potential noisy samples are detailed in Procedure~\ref{proc:label_relevance}. Crucially, both detection decisions and final label corrections are performed entirely at the user side. As a result, the PS never accesses the users’ true labels, preserving data privacy.

In summary, our approach exploits similarities between user distributions while focusing on subsets of samples that share a specific label, some of which may be mislabeled. In the clustering problem, we model each user's data as a mixture of different underlying distributions. Users with similar mixtures could be identified by comparing the eigenvalues of their Gram/covariance matrices with those estimated from projecting one user’s eigenvectors onto another user’s data. 

For the detection and correction challenge, we narrow the focus to individual components of this mixture, i.e., samples from each class considered separately. When samples are correctly labeled, their similarity with the corresponding public samples (clean ones available at the PS) is high, since their Gram matrices, which estimates the covariance of the class distribution, are nearly identical. In contrast, mislabeled samples yield covariance estimates that diverge from the true public distribution, leading to noticeable differences in the eigenvalues. To relabel such samples, we assign them to the class whose subspace provides the strongest representation of the sample under investigation.
Overall, both proposed approaches are strictly feature-centric and fully decoupled, offering significant communication-efficiency gain over model-based approaches.

\begin{algorithm}[t]
\caption{\texttt{FB-NLL}}\label{alg:main_algo}
\begin{algorithmic}[1]

\State \textbf{Input}: number of clusters $M$, users $\mathcal{K}$, global rounds $G$, local epochs $E$

\Statex \underline{\textit{(Cluster Identity Estimation)}}

\Statex \textbf{Step 1: Local spectral decomposition}
\For {$k \in \mathcal{K}$}
    \State Compute ${\bm{\Sigma}}_k = \frac{1}{n_k}\Phi(\mathbf{X}_k)^{\top}\Phi(\mathbf{X}_k)$
    \State Obtain top-$q$ eigenpairs $(\bm{\lambda}_k, \mathbf{V}_k)$
\EndFor

\Statex \textbf{Step 2: Cross-user similarity evaluation}
\For {$k \in \mathcal{K}$}
    \For {$j \in \mathcal{K}\setminus\{k\}$}
        \State User $j$ shares $\mathbf{V}_j$
        \State User $k$ computes $r(k,j)$ via Procedure~\ref{proc:data_relevance_general}
        \State User $k$ sends $r(k,j)$ to PS
    \EndFor
\EndFor

\Statex \textbf{Step 3: Symmetric similarity and clustering}
\State PS computes $[\mathbf{R}]_{k,j} = \frac{r(k,j)+r(j,k)}{2}$
\State Apply HAC on $\mathbf{R}$ to obtain cluster identity matrix $\bm{CI}$

\Statex \underline{\textit{(Noise Detection and Correction)}}
\State Execute Procedure~\ref{proc:label_relevance}

\Statex \underline{\textit{(Cluster-wise Training)}}
\State PS initializes $\{w_m^{(0)}\}_{m=1}^M$

\For {$t = 0,1,\dots,G-1$}
    \For {$k \in \mathcal{K}$}
        \State Let $m_k$ be such that $\bm{CI}[k,m_k]=1$
        \State $w_{(k,m_k)}^{(t)} \gets$ Local SGD on cluster model $w_{m_k}^{(t)}$ for $E$ epochs
    \EndFor
    
    \Statex \hspace{0.3in}\underline{\textit{(PS aggregation)}}
    \For {$m \in [M]$}
        \State 
        \[
        w_m^{(t+1)} =
        \frac{\sum_{k=1}^K \bm{CI}[k,m]\, w_{(k,m)}^{(t)}}
        {\sum_{k=1}^K \bm{CI}[k,m]}
        \]
    \EndFor
\EndFor

\end{algorithmic}
\end{algorithm}

\begin{procedure}[t]
\caption{Noisy Label Detection and Correction}
\label{proc:label_relevance}
\begin{algorithmic}[1]

\State PS shares class-specific principal subspaces $\{\mathbf{Q}_l^{c}\}_{c=1}^{C}$ with all users

\For{$k \in \mathcal{K}$}
    \For{each local class $\tilde{c}$ at user $k$}

        \State Extract samples with label $\tilde{c}$: $\mathbf{X}^{\tilde{c}}$
        \State Compute eigenvectors $\mathbf{V}^{\tilde{c}}$ of $\frac{1}{n^{\tilde{c}}}\Phi(\mathbf{X}^{\tilde{c}})^\top \Phi(\mathbf{X}^{\tilde{c}})$
        \State Upload $\mathbf{V}^{\tilde{c}}$ to PS

        \State Initialize similarity list $\mathbf{r}_\text{avg} = [\,]$

        \For{each clean class $c \in [C]$}
            \State PS computes $r(c,\tilde{c})$ using Procedure~\ref{proc:data_relevance_general}
            \State PS sends $r(c,\tilde{c})$ to user $k$
            \State User computes reciprocal similarity $r(\tilde{c},c)$
            \State Append $\frac{r(c,\tilde{c}) + r(\tilde{c},c)}{2}$ to $\mathbf{r}_\text{avg}$
        \EndFor

        \State $\mathcal{I} \gets \{ c \in [C] : \mathbf{r}_\text{avg}[c] \ge \tau_{\text{sim}} \}$

        \If{$|\mathcal{I}| = 1$}
            \State Relabel entire class $\tilde{c}$ to $c^\star \in \mathcal{I}$
        \Else
            \For{each feature $\mathbf{z} \in \Phi(\mathbf{X}^{\tilde{c}})$}
                \State $\tilde{c} \leftarrow \arg\max_{c \in [C]} 
                \left\| \mathbf{Q}_l^c (\mathbf{Q}_l^c)^\top \mathbf{z} \right\|_2$
            \EndFor
        \EndIf

    \EndFor
\EndFor

\end{algorithmic}
\end{procedure}

\section{Experiments} \label{sec:experiments}

\textbf{Datasets and Models.}
We validate the proposed $\texttt{FB-NLL}$ framework and all baselines on multiple datasets, model architectures, and task configurations. Specifically, we consider ResNet-based models \cite{He_2016_CVPR}, using ResNet18 on CIFAR-10 \cite{krizhevsky2009learning} and ResNet10 on SVHN \cite{svhn}.

We set the number of users to $K=20$ and partition them equally among the predefined tasks. To satisfy the clean-validation assumption at the PS, we sample a small, class-balanced subset of $n_s^c$ examples per class $c$ to construct a clean validation set. This dataset is disjoint from all users’ local data and is used solely to compute class-specific principal subspaces during the noise-correction stage. We set $n_s^c=200$ for CIFAR-10 and $n_s^c =300$ for SVHN, corresponding to approximately $4\%$ of the total training data. 

The remaining samples are assigned to users according to task definitions, where each task corresponds to a distinct subset of classes. For each task, its associated training samples are distributed among the users of that task in an independent and identically distributed (IID) manner. Consequently, although data is approximately IID within each task group, the global data distribution across users is highly heterogeneous. This setting challenges single-model federated learning and highlights the necessity of accurate cluster identity estimation.

To evaluate robustness under cross-task contamination, we introduce data impurity. For each task group, $8\%$ of its training samples are randomly selected and redistributed uniformly across all users, irrespective of their assigned tasks.

Unless otherwise stated, we use a learning rate of $5\times 10^{-4}$, train for $2$ epochs per communication round with a batch size of $64$, and employ SGD with momentum $0.5$ as the local optimizer. Training is conducted for $80$ communication rounds. To mitigate the effect of noisy labels during training, we set the weight decay parameter to $0.001$.

\textbf{Tasks.} We define the task configurations as summarized in Table~\ref{tab:tasks}. In the two-task setting, the partitions represent distinct semantic splits: vehicle versus non-vehicle classes for CIFAR-10 and odd versus even digits for SVHN. For the three- and five-task settings, we construct alternative label groupings that partition the class space into multiple disjoint subsets. These configurations simulate diverse user preferences and increasing levels of personalization granularity. These configurations show that the proposed framework does not rely on any specific task structure. 

\textbf{Noise Models.} We evaluate our proposed algorithm under three label corruption models: class-independent, class-dependent, and uniform noise. For the class-independent and class-dependent settings, the sample-level noise rate is set to $25\%$. For the uniform setting, we set the fraction of noisy users to $\rho = 0.4$ and the lower bound of the user-specific noise rate to $\beta = 0.2$, thereby simulating high-noise scenarios.

\textbf{Feature Mapping Function.}\label{sec:feature_mapping function}
The feature mapping function $\Phi$ plays a central role in our framework by extracting informative embeddings from raw data and reducing their dimensionality, thereby mitigating dimensional complexity.
The quality of these embeddings directly impacts both clustering and noisy-label correction stages. 
Various approaches can be used to construct $\Phi$. A common strategy is to extract representations using a pre-trained deep model \cite{data_valuation, Many-taskFL}. Alternatively, the Histogram of Oriented Gradients (HoG) \cite{HoG} provides lightweight and label-agnostic feature representations, making it inherently robust to noisy labels. In \cite{RCC_icc25}, we demonstrate the effectiveness of the HoG approach for clustering across different datasets.

In this work, to highlight the flexibility of our proposed framework, we employ diverse feature mappings depending on the dataset and stage. 
For the clustering stage, we use HoG features for CIFAR-10 to obtain lightweight geometric representations, while for SVHN we extract features using a pre-trained ResNet-18 model. In contrast, the noisy-label detection and correction stage requires finer semantic discrimination to accurately separate class-specific subspaces. Therefore, we employ pre-trained ResNet-18 features for this stage across datasets to obtain richer and more semantically meaningful representations. This stage-dependent design underscores the modularity of the framework, which remains agnostic to the specific choice of feature mapping as long as meaningful representations are provided. It is worth noting that all FL models are initialized randomly and trained from scratch.

\textbf{Rank Selection.} We determine the truncation rank $q$ adaptively by retaining dominant spectral components whose eigenvalues exceed a small threshold, which is fixed per dataset. This follows standard spectral practice of discarding low-energy directions that are often dominated by noise \cite{data_valuation}.

Furthermore, in the detection and correction stage, we allow different truncation ranks in Phases~1 and~2. Phase~1 compares class-level covariance structures and therefore relies primarily on dominant spectral components. In contrast, Phase~2 performs sample-level relabeling via subspace projection and benefits from a slightly richer class subspace representation to better capture intra-class variability. Accordingly, the projection rank in Phase~2 may differ from that in Phase~1.

\subsection{Clustering Methodology Evaluation (\texttt{FB-NLL(-)})}
We first evaluate the effectiveness of our feature-based clustering methodology for cluster identity estimation in PFL. To isolate the clustering component and assess its robustness to noisy labels, we apply Algorithm~\ref{alg:main_algo} without the noise detection and correction Procedure~\ref{proc:label_relevance}. This ablated variant, denoted as \texttt{FB-NLL(-)}, is compared with multiple baselines in Tables~\ref{tab:class_indep_cluster}, \ref{tab:class_dep_cluster}, and \ref{tab:noise_type3_cluster}.

\textbf{Baselines.} We consider two baseline algorithms.

The first baseline is an {\it iterative} clustering scheme, denoted {\it IFCA-PFL}, which is adapted from the Iterative Federated Clustering Algorithm (IFCA) \cite{clustered_FL} to our PFL setting. It assigns users to clusters based on their loss values. Specifically, user $k$ is assigned to cluster
\[
m_k = \arg\min_{m\in[M]} F_k(w_m^{(t)}),
\]
where $F_k(w)$ denotes the empirical loss of model $w$ evaluated on user $k$'s local dataset. Since vanilla IFCA \cite{clustered_FL}  may converge to degenerate solutions in which all users collapse into a single cluster, we modify it by enforcing a predefined number of clusters at each global communication round. This modification guarantees exactly $M$ active clusters throughout training and ensures stable personalized learning. The complete procedure is summarized in Algorithm~\ref{alg:iter_algo}.

The second baseline is a {\it single-group} scheme, corresponding to the case where no clustering is performed and all users collaboratively train a single global model (i.e., standard FL setting). This baseline highlights the significance of clustering under heterogeneous task distributions and is included to motivate the necessity of clustering especially under heterogeneous task distributions.

\begin{algorithm}[t]
\caption{IFCA-PFL}\label{alg:iter_algo}
\begin{algorithmic}[1]
\State \textbf{Input}: number of tasks $M$, number of global iterations $G$, number of users $K$, number of local epochs $E$
\State $(\bm{CI}, \{w_m^{(0)}\}_{m=1}^M) \leftarrow$ Execute Procedure~\ref{proc:ifca_init}
\State PS broadcasts $\{w_m^{(0)}\}_{m=1}^M$

\For {$t = 0, \dots, G-1$}
    \For{$k \in \mathcal{K}$}
        \State Let $m_k$ be such that $\bm{CI}[k,m_k]=1$
        \State $w_{k,m_k}^{(t)} \leftarrow$ Perform SGD on $w_{m_k}^{(t)}$ for $E$ epochs
        \State User $k$ uploads $w_{(k,m_k)}^{(t)}$ to PS
    \EndFor

    \Statex \hspace{0.25in}\underline{\textit{(PS aggregation)}}
    \For{$m = 1, \dots, M$}
        \State $w_m^{(t+1)} =
        \frac{1}{\sum_{k=1}^{K} \bm{CI}[k,m]}
        \sum_{k=1}^{K} \bm{CI}[k,m]\, w_{(k,m)}^{(t)}$
    \EndFor

    \State $N_m \gets \sum_{k=1}^{K}\bm{CI}[k,m],\ \forall m\in[M]$
    \State $\bm{CI} \leftarrow$  Procedure~\ref{proc:group_reassociation}$(\{w_m^{(t+1)}\}_{m=1}^M, \mathcal{K}, \{N_m\}_{m=1}^M)$
\EndFor
\end{algorithmic}
\end{algorithm}

\begin{procedure}[t]
\caption{IFCA-PFL Initialization (Nonempty Clusters)}\label{proc:ifca_init}
\begin{algorithmic}[1]
\State \textbf{Input}: number of clusters $M$, user set $\mathcal{K}$, local datasets $\{\mathcal{D}_k\}_{k\in\mathcal{K}}$
\State \textbf{Output}: initial cluster assignment matrix $\bm{CI}$, initial models $\{w_m^{(0)}\}_{m=1}^M$

\Repeat
    \State Initialize $M$ models $\{w_m^{(0)}\}_{m=1}^M$ randomly
    \State $\bm{CI} \gets \mathbf{0}_{K\times M}$
    \For{$k \in \mathcal{K}$}
        \State $m_k \gets \arg\min_{m\in[M]} F_k\!\left(w_m^{(0)}\right)$
        \State $\bm{CI}[k,m_k] \gets 1$
    \EndFor
\Until{$\sum_{k=1}^K \bm{CI}[k,m] \ge 1,\ \forall m\in[M]$}
\State \Return $\bm{CI},\{w_m^{(0)}\}_{m=1}^M$
\end{algorithmic}
\end{procedure}


\begin{procedure}[t]
\caption{Cluster Re-association}\label{proc:group_reassociation}
\begin{algorithmic}[1]
\State \textbf{Input}: models $\{w_m^{(t)}\}_{m=1}^M$, user set $\mathcal{K}$, previous cluster sizes $\{N_m\}_{m=1}^M$
\State \textbf{Output}: updated cluster assignment matrix $\bm{CI}$

\State Initialize loss matrix $\mathbf{L}\in\mathbb{R}^{K\times M}$
\For{$k\in\mathcal{K}$}
    \For{$m\in[M]$}
        \State $\mathbf{L}[k,m] \gets F_k\!\left(w_m^{(t)}\right)$
    \EndFor
\EndFor

\Statex \underline{\textit{(loss-minimizing assignment)}}
\State Initialize $\bm{CI}\gets \mathbf{0}_{K\times M}$
\For{$k\in\mathcal{K}$}
    \State $\hat{m}_k \gets \arg\min_{m\in[M]} \mathbf{L}[k,m]$
    \State $\bm{CI}[k,\hat{m}_k]\gets 1$
\EndFor
\Statex \underline{\textit{(PS-side)}}
\If{$\sum_{k=1}^K \bm{CI}[k,m] \ge 1,\ \forall m\in[M]$}
    \State \Return $\bm{CI}$
\EndIf

\State Initialize $\bm{CI}\gets \mathbf{0}_{K\times M}$

\State $\tilde{\mathbf{L}} \gets \mathbf{L}$ 
\For{$m=1,\dots,M$}
    \State Select $\mathcal{S}_m \gets$ the set of indices of the $N_m$ smallest entries in $\tilde{\mathbf{L}}[:,m]$
    \For{$k \in \mathcal{S}_m$}
        \State $\bm{CI}[k,m]\gets 1$
    \EndFor
    \State Set $\tilde{\mathbf{L}}[k,:]\gets +\infty,\ \forall k\in\mathcal{S}_m$ \Comment{remove selected users}
\EndFor
\State \Return $\bm{CI}$
\end{algorithmic}
\end{procedure}

\begin{table}[t]
 \begin{center}
    \caption{Task definitions.}
    \label{tab:tasks}
   \resizebox{\columnwidth}{!} { \begin{tabular}{|c|c|c|c|}
    \hline
      \textbf{Dataset}  & \textbf{Two Tasks}&\textbf{Three Tasks} & \textbf{Five Tasks}\\
      \hline

    \multirow{3}{*}{CIFAR-10} & $T_1=\{0,1,8,9\}$ &   $T_1=\{0,1,8\}$&   $T_1=\{0,1\},~ T_2=\{8,9\} $ \\ 
      & $T_2=\{2,3,4,5,6,7\}$&  $T_2=\{9,2,3\}$ & $T_3=\{2,3\},~ T_4=\{4,5\}$ \\ 
     & & $T_3=\{4,5,6,7\}$ & $T_5=\{6,7\}$ \\

      \hline
            \multirow{3}{*}{SVHN} & $T_1=\{0,2,4,6,8\}$ &   $T_1=\{0,2,4\}$&   $T_1=\{0,2\},~ T_2=\{4,6\} $ \\ 
      & $T_2=\{1,3,5,7,9\}$&  $T_2=\{6,8,1\}$ & $T_3=\{8,1\},~ T_4=\{3,5\}$ \\ 
     & & $T_3=\{3,5,7,9\}$ & $T_5=\{7,9\}$ \\

      \hline
    \end{tabular}}
  \end{center}
\end{table}

\begin{table}[t]
 \begin{center}
    \caption{Clustering accuracy under class-independent noise.}
    \label{tab:class_indep_cluster}
   \resizebox{\columnwidth}{!} { \begin{tabular}{|c|c|c|c|c|}
    \hline
      \textbf{Dataset} & \textbf{Algorithms} & \textbf{Two Tasks}&\textbf{Three Tasks} & \textbf{Five Tasks}\\
      \hline

    \multirow{4}{*}{CIFAR-10}

       & FB-NLL(-) & \bm{$71.58 \pm 0.43$} & \bm{  $83.25 \pm 0.27$} & \bm{$90.99 \pm 0.14$}\\ 
       & IFCA-PFL  & $67.93 \pm 3.97$
 & $80.03 \pm 6.88$& $89.43 \pm 3.30$ \\
       & Single global model & $56.61 \pm 1.21$  & $52.22 \pm 0.65$& $47.96 \pm 0.79
$ \\
\hline
\multirow{4}{*}{SVHN}
       & FB-NLL(-) & $\bm{86.35 \pm 0.59}$ & $\bm{  90.51 \pm 0.33 }$ & $\bm{ 94.89 \pm 0.17}$\\ 
       & IFCA-PFL  & $67.05 \pm 5.75$
 & $85.89 \pm 6.68$&  $87.08 \pm 5.22$ \\
       & Single global model & $ 65.71 \pm 1.62 $ & $60.83 \pm 2.57$& $ 39.80 \pm 4.57$   \\
\hline
    \end{tabular}}
  \end{center}
\end{table}

\begin{table}[t]
 \begin{center}
    \caption{Clustering accuracy under class-dependent noise.}
    \label{tab:class_dep_cluster}
   \resizebox{\columnwidth}{!} { \begin{tabular}{|c|c|c|c|c|}
    \hline
      \textbf{Dataset} & \textbf{Algorithms} & \textbf{Two Tasks}&\textbf{Three Tasks} & \textbf{Five Tasks}\\
      \hline

    \multirow{4}{*}{CIFAR-10}
      & FB-NLL(-) & \bm{$72.19 \pm 1.21$}  & \bm{$83.54 \pm 1.11 $} & \bm{$91.73 \pm 0.24$ }\\ 
      & IFCA-PFL  & $70.13 \pm 4.35$ & $69.86 \pm 14.2$ & $78.77 \pm 13.1$\\
      & Single global model &  $58.46 \pm 1.50$  & $53.21 \pm 1.86$ & $47.77 \pm 1.65$ \\
\hline
    \multirow{4}{*}{SVHN}
      & FB-NLL(-) & \bm{$ 85.30 \pm 1.62$}  & \bm{$89.74 \pm 1.58 $} & \bm{$ 93.17 \pm 1.97$ }\\ 
      & IFCA-PFL  & $ 80.19 \pm 10.6 $ & $80.58 \pm 9.38 $ & $82.58 \pm 6.45 $\\
      & Single global model &  $67.60 \pm 2.21$  & $61.95 \pm 2.55 $ & $42.77 \pm  4.05$ \\
\hline
    \end{tabular}}
  \end{center}
\end{table}

\begin{table}[t]
 \begin{center}
    \caption{Clustering accuracy under uniform noise.}
    \label{tab:noise_type3_cluster}
   \resizebox{\columnwidth}{!} { \begin{tabular}{|c|c|c|c|c|}
    \hline
      \textbf{Dataset} & \textbf{Algorithms} & \textbf{Two Tasks}&\textbf{Three Tasks} & \textbf{Five Tasks}\\
      \hline
    \multirow{4}{*}{CIFAR-10}
      & FB-NLL(-) & \bm{$69.69 \pm 2.5$}  & \bm{$80.38 \pm 2.28$} & \bm{$89.83 \pm 1.05$} \\ 
      & IFCA-PFL  & $67.94 \pm 4.01$  & $76.50 \pm 5.41$ &  $80.65 \pm 3.01$\\
      & Single global model & $55.73 \pm 2.76$   & $51.23 \pm 2.26$  & $48.44 \pm 2.10$ \\
\hline
    \multirow{4}{*}{SVHN}
      & FB-NLL(-) & $\bm{85.25 \pm 2.20}$ &  $\bm{88.86 \pm 2.13}$ & $\bm{93.83 \pm 0.80}$ \\ 
      & IFCA-PFL  &  $85.17 \pm 1.64$ & $83.96 \pm 8.65$ & $88.43 \pm 8.51$ \\
      & Single global model & $64.05 \pm 1.81 $ & $59.92 \pm 2.24$ & $ 40.74 \pm 4.16$ \\
\hline
    \end{tabular}}
  \end{center}
\end{table}


\textbf{Beating the Baselines.} Tables~\ref{tab:class_indep_cluster}--\ref{tab:noise_type3_cluster} report the average accuracy and standard deviation over five runs. Across all datasets, noise types, and task configurations, \texttt{FB-NLL(-)} consistently outperforms both IFCA-PFL and the single global model, and remains robust under both task-dependent (class-independent and class-dependent) and task-independent (uniform) noise, with only minor performance variations across noise types. 
Notably, performance improves as the number of tasks increases from two to five. We attribute this to the reduced number of intended learning labels per task in larger-task settings (see Table~ \ref{tab:tasks}).

In contrast, IFCA-PFL deteriorates sharply, particularly under class-dependent noise. Since  IFCA-PFL relies on empirical loss values for cluster assignment, structured label noise can distort loss estimates and lead to wrong clustering decisions. This is also reflected in its substantially high variance across runs. 
The results further indicate that the single global model performs significantly worse across all configurations, especially in the five-task setting, and confirms that learning a single shared model under heterogeneous task distributions is suboptimal and underscores the necessity of effective clustering in PFL. 
 
\noindent \textbf{Dataset/Distribution Robustness.} We further evaluate the robustness of the proposed clustering method under heterogeneous data distributions drawn from different datasets. In particular, we consider a three-user setting involving both CIFAR-10 and CIFAR-100.  User 1 has vehicles-related labels from CIFAR-10, while users $2$ and $3$ have CIFAR-100 data, with user $2$ having labels representing vehicles, and user $3$ having labels from the remaining classes of CIFAR-100.

The results, as shown in Table~\ref{tab:cifar10-cifar100}, show the efficacy of our clustering algorithm in successfully matching users with similar labels, even when such labels originate from different datasets. This showcases the framework's ability to handle diverse label distributions and facilitate effective collaboration among users.

\begin{figure}[t]
    \centering
    \includegraphics[width=.8\linewidth]{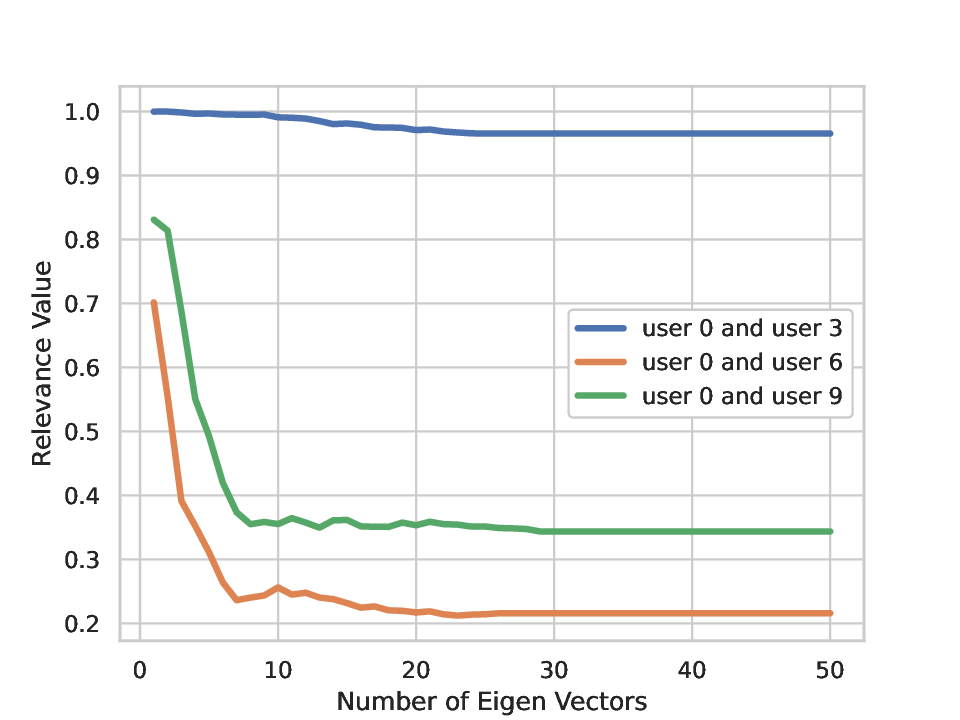}
    \caption{Sensitivity of the relevance value to eigenvector dimensionality. Users $0$ and $3$ share the same task, whereas users $6$ and $9$ belong to different tasks.}
    \label{fig:n_eigens}
\end{figure}

\begin{table}[h]
\centering
\begin{tabular}{c|cc}
\hline
Users & User $2$ & User $3$ \\
\hline
User $1$ & $0.62$ & $0.39$ \\
\hline
\end{tabular}
\caption{Similarity score between users having different datasets: CIFAR-10 (User $1$) and CIFAR-100 (Users $2$ and $3$).}
\label{tab:cifar10-cifar100}
\vspace{-.1in}
\end{table}
\noindent \textbf{Communication Improvement.}  To illustrate the communication efficiency of the proposed clustering algorithm, Fig.~\ref{fig:n_eigens} shows that effective clustering can be achieved using as few as ten eigenvectors. Following the experimental setting of \cite{ds_asilomar24}, we select the top $10$ eigenvectors corresponding to the largest eigenvalues. This implies that instead of exchanging the full matrix $\mathbf{V} \in \mathbb{R}^{784 \times 784}$, users can transmit only a truncated matrix $\mathbf{V}\in \mathbb{R}^{784 \times 10}$ and still accurately estimate pairwise relatedness.

This finding emphasizes a key advantage of the proposed feature-based approach over weight-based clustering methods, which typically require exchanging full model parameters with dimensionality on the order of millions. By contrast, our methods requires only a small number of principal directions, making it substantially more communication-efficient.

\subsection{Detection and Correction Methodology Evaluation}

While the proposed clustering mechanism is independent of label correctness, noisy labels can still degrade the downstream learning performance as shown in Table~\ref{tab:clean_label_perf} compared to the achieved performances under \texttt{FB-NLL(-)}. Therefore, we now evaluate the noise detection and correction component, which mitigates the effects of label corruption within each identified cluster prior to federated optimization. 

\begin{table}[t]
 \begin{center}
    \caption{Clean-label Performance (no noise).}
    \label{tab:clean_label_perf}
   \resizebox{\columnwidth}{!} { \begin{tabular}{|c|c|c|c|}
    \hline
      \textbf{Dataset}  & \textbf{Two Tasks}&\textbf{Three Tasks} & \textbf{Five Tasks}\\
      \hline
   {CIFAR-10}
      & $76.00 \pm 0.50$ & $ 86.06 \pm  0.25$& $ 92.39 \pm 0.23 $ \\
\hline
   {SVHN}
   & $ 90.99 \pm 0.20 $ &  $ 93.71 \pm 0.32$ & $ 96.49 \pm 0.16$ \\ 
      
\hline
    \end{tabular}}
  \end{center}
 
\end{table}

The integration of feature-based clustering and noise correction constitutes the full \texttt{FB-NLL} framework.  

\textbf{Baselines.} We consider three baseline algorithms.

The first baseline is the {\it FedCorr} algorithm \cite{FedCorr}, a multi-stage framework for noisy-label correction in FL. In the first stage, users are identified as clean or noisy based on their accumulated local intrinsic dimension scores using a Gaussian Mixture Model (GMM). In the second stage, clean users assist noisy users in relabeling their potentially mislabeled samples. Finally, all users participate in a global fine-tuning phase. We set the first-stage to $2$ rounds. FedCorr adopts the fraction scheduling strategy, where each user participates exactly once per iteration. Under this strategy, one iteration requires $K$ communication rounds, and thus the first stage consumes $2K$ rounds in total. The second stage, label correction for noisy users combined with federated averaging over clean users, is executed for $10$ rounds. The remaining rounds are allocated to standard FL. All other hyperparameters follow the original FedCorr configuration. 

The second baseline is the \textit{FedClip} algorithm \cite{ClientPruning}, a multi-stage framework consisting of two stages. In the first stage, FedClip identifies clean and noisy users based on their performance on a clean validation dataset at the PS.\footnote{For fair comparison, the same clean validation dataset at the PS is used across all methods.} Users with the lowest validation performance are pruned, and FL proceeds using only the top-performing fraction of users. We set the number of rounds in the first stage to $20$. Among the users identified as clean, half are selected to participate in the subsequent fine-tuning stage.

The third baseline is the \textit{RHFL} algorithm \cite{fang2022robust}, which mitigates the impact of noisy labels through a noise-robust loss formulation. Specifically, RHFL symmetrically combines the standard cross-entropy loss with the reverse cross-entropy loss to alleviate overfitting to mislabeled samples. In addition, training is regularized by aligning users' logit distributions using a clean public dataset available at the PS.
 
\textbf{Beating the Baselines.} 
Across all noise types, datasets, and task configurations, \texttt{FB-NLL} achieves the highest accuracy with consistently low variance, as shown in Tables~\ref{tab:class_indep_corr}--\ref{tab:class_uniform_corr}. These gains remain stable as the number of tasks increases, demonstrating effective scaling with task complexity. 

\texttt{FB-NLL} exhibits superior performance under class-dependent noise due to its ability to exploit structural alignment. In the class-dependent setting, corrupted samples from a given class are systematically mapped to a specific incorrect label, creating a coherent displacement in the feature space. When these samples are compared against the PS's clean-specific principal subspaces, this structured deviation becomes detectable through the similarity metric defined in \eqref{eq:eigen_proj}--\eqref{eq:avg_r}. Since the metric evaluates alignment across the dominant spectral components, mismatched subspaces incur a similarity penalty, enabling \texttt{FB-NLL} to effectively distinguish wrong labels from their true counterparts and restore label consistency. In contrast, loss-based methods are more vulnerable under class-dependent noise because the corruption is systematic and may produce low training loss despite semantic misalignment.

As for uniform and class-independent noise, corruption is distributed across all labels, making correction more difficult for baselines that depend on global loss statistics or pruning: Fedcorr underscores the importance of incorporating label correction in noisy-label learning approaches; 
ClipFL performs competitively under class-independent and uniform noise but degrades under class-dependent noise, reflecting the risk of pruning users who possess exclusive task-specific labels; and, RHFL’s performance highlights the limitations of robustness methods that depend solely on loss functions in federated settings with heterogeneous and substantial label noise.

\textbf{Modularity.} One defining advantage of the \texttt{FB-NLL} framework is its \textit{modular} design, which enables it to function as a plug-and-play enhancement for existing FL baselines. As demonstrated in our empirical evaluation, hybrid configurations---such as FedCorr+FB-NLL, ClipFL+FB-NLL, and RHFL+FB-NLL--- generally achieve higher accuracy and lower variance across diverse noise models and task configurations. In particular, integration with RHFL yields substantial gains, with absolute accuracy improvements ranging from $3.5$ to $24$ percentage points. Similarly, ClipFL benefits from improvements up to $10$ percentage points. While FedCorr also shows gains (typically between $0.05$ to $4.8$ percentage points), performance is sensitive to the integration point within its multi-stage pipeline. For example, in the SVHN five-task setting under class-independent noise, applying \texttt{FB-NLL} immediately after FedCorr's initial identification stage results in a minor accuracy decrease of $1.79$ percentage points. However, when integration is applied after FedCorr's second stage---thereby avoiding interference with its internal correction mechanism---the accuracy increases by $0.24$ percentage points. 
These results indicate that while \texttt{FB-NLL} is highly adaptable, careful selection of the integration stage is important to preserve the training dynamics of the underlying baseline. 

It is important to note that integrating \texttt{FB-NLL} does not replace the core label-correction or communication mechanisms of the baseline methods. Consequently, the final performance remains influenced by the inherent training dynamics of the underlying algorithms. This interaction explains why the hybrid configurations, although significantly improved, {\it do not surpass the performance of the standalone \texttt{FB-NLL} framework}.


\begin{table}[t]
 \begin{center}
    \caption{PFL performance under class-independent noise across various label detection and correction methods.}
    \label{tab:class_indep_corr}
   \resizebox{\columnwidth}{!} { \begin{tabular}{|c|c|c|c|c|}
    \hline
      \textbf{Dataset} & \textbf{Algorithms} & \textbf{Two Tasks}&\textbf{Three Tasks} & \textbf{Five Tasks}\\
      \hline
    \multirow{4}{*}[-3.3em]{\centering CIFAR-10} 
      & FB-NLL  & \bm{$74.77 \pm 0.46$} & \bm{ $85.13 \pm 0.2 $} & \bm{ $91.96 \pm 0.18$}\\ 
      & FedCorr & $70.85 \pm 0.56$  & $82.81 \pm 0.47$ &  $90.01 \pm 0.69$\\ 
      
      &ClipFL &  $70.70 \pm 0.44 $ & $80.80 \pm 0.36$ &  $88.30 \pm 0.80 $\\
      &RHFL & $49.32 \pm 1.50 $  & $59.72 \pm 1.85 $ & $70.42 \pm 2.14$\\ [3pt]
\cline{2-5} 
      & FedCorr + FB-NLL & $ 72.91 \pm 0.83 $ & $ 84.1 \pm 0.74 $ & $ 90.75 \pm 0.48$ \\
      & ClipFL + FB-NLL  & $73.46 \pm  0.67$ & $82.85 \pm 0.46 $ & $90.16 \pm 0.23$  \\
      & RHFL + FB-NLL    & $64.99 \pm  0.37 $ & $77.20 \pm 0.76 $ & $87.37 \pm 0.28$ \\
    \hline
    \multirow{4}{*}[-3.3em]{\centering SVHN} 
  & FB-NLL  & \bm{$ 88.42 \pm 0.36$} & \bm{$92.07 \pm 0.20$} & \bm{$ 94.94 \pm 0.70$} \\ 
  & FedCorr & $84.10 \pm 0.62 $ & $89.06 \pm 1.55$ & $93.57 \pm 0.30$ \\ 
  & ClipFL  & $ 85.92 \pm 0.81$  & $89.46 \pm 0.50$ & $92.88 \pm 1.28$ \\
  & RHFL    & $66.50 \pm 0.96$ & $72.34 \pm 0.69$ & $ 80.07 \pm 1.38$ \\ [3pt]
\cline{2-5} 
  & FedCorr + FB-NLL & $85.26 \pm 0.66$ & $89.92 \pm 1.50 $ & $91.78 \pm 2.61$ \\
  & ClipFL + FB-NLL  & $87.27 \pm 0.56$ & $90.77 \pm 0.51$ & $ 93.64 \pm 0.23$ \\
  & RHFL + FB-NLL    & $81.45 \pm 0.63$ & $85.72 \pm 0.59$ & $89.84 \pm 1.02$ \\
\hline
      
    \end{tabular}}
  \end{center}
\end{table}

\begin{table}[t]
 \begin{center}
    \caption{PFL performance under class-dependent noise across various label detection and correction methods.}
    \label{tab:class_dep_corr}
   \resizebox{\columnwidth}{!} { \begin{tabular}{|c|c|c|c|c|}
    \hline
      \textbf{Dataset} & \textbf{Algorithms} & \textbf{Two Tasks}&\textbf{Three Tasks} & \textbf{Five Tasks}\\
      \hline
    \multirow{4}{*}[-3.3em]{\centering CIFAR-10} 
      & FB-NLL  & \bm{$75.92 \pm 0.49 $} & \bm{ $ 86.16\pm  0.18 $} & \bm{ $ 92.44 \pm 0.16 $}\\ 
      & FedCorr & $ 70.59 \pm 2.02 $  & $ 82.42 \pm 0.83 $ &  $ 90.53 \pm 0.72 $\\ 
      
      &ClipFL &  $ 67.57 \pm 3.05 $ & $ 73.36 \pm 2.61 $ &  $ 87.57 \pm  2.24$\\
      &RHFL & $48.18  \pm 0.27 $  & $ 56.56 \pm  0.60 $ & $ 64.92 \pm 1.73 $\\ [3pt]
\cline{2-5} 
      & FedCorr + FB-NLL & $ 72.96 \pm 0.49 $ & $  84.35 \pm 0.56 $ & $ 91.07 \pm 0.36 $ \\
      & ClipFL + FB-NLL  & $74.37  \pm  0.78 $ & $83.71 \pm 0.30 $ & $ 90.52 \pm 0.14$  \\
      & RHFL + FB-NLL    & $65.59 \pm  0.80 $ & $ 79.02 \pm 0.43$ & $ 89.25 \pm 0.48$ \\
    \hline
      
\multirow{4}{*}[-3.3em]{\centering SVHN} 
  & FB-NLL  & \bm{$90.28 \pm 0.20 $} & \bm{$ 93.43 \pm 0.32$} & \bm{$96.41 \pm 0.18$} \\ 
  & FedCorr & $82.08 \pm 2.00$  & $86.37 \pm 2.62$ & $89.11 \pm 3.54$ \\ 
  & ClipFL  & $80.18 \pm 4.01$ & $83.37 \pm 2.03$ & $93.43 \pm1.66$ \\
  & RHFL    & $63.53 \pm 1.12$ & $66.93 \pm 1.20$ & $ 70.01 \pm 1.65$ \\ [3pt]
\cline{2-5} 
  & FedCorr + FB-NLL & $86.89 \pm 1.36$ & $90.02 \pm 1.29$ & $92.38 \pm 0.90$ \\
  & ClipFL + FB-NLL  & $89.74 \pm 0.28$ & $92.56 \pm 0.29$ & $95.54 \pm 0.19 $ \\
  & RHFL + FB-NLL    & $84.78 \pm 0.64$ & $89.01 \pm 0.19 $ & $93.99 \pm 0.46$ \\
\hline
    \end{tabular}}
  \end{center}
\end{table}

\begin{table}[t]
 \begin{center}
    \caption{PFL performance under uniform noise across various label detection and correction methods.}
    \label{tab:class_uniform_corr}
   \resizebox{\columnwidth}{!} { \begin{tabular}{|c|c|c|c|c|}
    \hline
      \textbf{Dataset} & \textbf{Algorithms} & \textbf{Two Tasks}&\textbf{Three Tasks} & \textbf{Five Tasks}\\
      \hline
    \multirow{4}{*}[-3.3em]{\centering CIFAR-10} 
      & FB-NLL  & \bm{$ 74.75 \pm 0.69$} & \bm{ $ 85.05 \pm  0.34 $} & \bm{ $ 91.77 \pm 0.49$}\\ 
      & FedCorr & $ 71.52 \pm 1.42$  & $ 82.90 \pm 1.01 $ &  $90.25 \pm 0.42 $\\ 
      
      &ClipFL &  $ 74.53 \pm 1.24$ & $ 83.23 \pm 0.95 $ &  $ 90.11 \pm 0.79 $\\
      &RHFL & $ 56.84 \pm 3.01 $  & $69.03 \pm 3.38 $ & $ 78.90 \pm 4.38 $\\ [3pt]
\cline{2-5} 
      & FedCorr + FB-NLL & $ 72.69  \pm 0.84 $ & $  83.11 \pm 0.89$ & $ 90.30 \pm 0.40 $ \\
      & ClipFL + FB-NLL  & $ 74.69 \pm 1.14 $ & $83.44 \pm 0.97 $ & $ 90.46 \pm 0.42$  \\
      & RHFL + FB-NLL    & $ 64.22 \pm  0.73 $ & $75.99 \pm 1.01$ & $86.37 \pm 1.55 $ \\
    \hline
    \multirow{4}{*}[-3.3em]{\centering SVHN} 
  & FB-NLL  & $\bm{87.79\pm 1.47}$ & $\bm{ 91.26 \pm 1.34}$ & $\bm{94.84 \pm 0.56}$ \\ 
  & FedCorr & $84.59 \pm 1.35$ & $87.92 \pm 1.87$ & $89.60 \pm 3.11$ \\ 
  & ClipFL  & $74.31 \pm 6.38$ & $ 76.02 \pm 6.27$ & $ 90.37 \pm 2.21$ \\
  & RHFL    &  $73.63 \pm 4.05$ & $77.97 \pm 3.70 $ & $85.91 \pm 2.47$ \\ [3pt]
\cline{2-5} 
  & FedCorr + FB-NLL & $85.28 \pm 1.51$ & $88.83 \pm 1.56$ & $ 90.02 \pm 2.92$ \\
  & ClipFL + FB-NLL  & $80.79 \pm 5.31$ & $ 83.39 \pm 3.77$ & $92.51 \pm 1.36$ \\
  & RHFL + FB-NLL    & $79.00 \pm 2.37$ & $ 83.44 \pm 2.20 $ & $89.46 \pm 1.60$ \\
\hline

    \end{tabular}}
  \end{center}
\end{table}



\section{Conclusion} \label{sec:conclusion}
We proposed \texttt{FB-NLL}, a holistic, \textit{feature-centric} framework that addresses two  major challenges in PFL: cluster-identity estimation and noisy-label detection and correction. The framework is communication-efficient and model-agnostic. Moreover, its two main components---cluster-identity estimation and noisy-label correction---are modular and can be seamlessly integrated into existing methods as plug-in modules. Extensive experiments have shown that \texttt{FB-NLL} consistently outperforms existing baselines across diverse noise models and heterogeneous task configurations.

\textbf{Future Directions.} Several extensions of the proposed framework merit further investigation. First, the framework may be generalized to support soft clustering, whereby each user is permitted to participate in multiple clusters with associated membership weights. Such an extension may better capture overlapping task structures and provide a systematic tradeoff between generalization and personalization. 

Furthermore, extending the framework beyond classification problems is of considerable interest. Another important direction is the study of non-IID data distributions within each task, where users may possess highly imbalanced local datasets. A particularly challenging setting arises when a user aims to learn a task for which only limited local samples (or none at all) are available. In such scenarios, feature \textit{diversity} across users may play a critical role. Designing algorithms that explicitly exploit cross-user feature diversity to support these users represents a promising direction for future research.

\bibliographystyle{unsrt}
\bibliography{Ali}
\end{document}